\documentclass[letterpaper, 10 pt, journal, twoside]{IEEEtran}
\usepackage{cite}

\ifCLASSINFOpdf
  \usepackage[pdftex]{graphicx}
\else
\fi
\usepackage{subcaption} 
\usepackage{subfig}
\usepackage{multirow}

\usepackage{tikz}

\begin{document}

%
\title{From Decision to Action in Surgical Autonomy: Multi-Modal Large Language Models for Robot-Assisted Blood Suction}
%
%
%

\author{Sadra Zargarzadeh$^{1}$,
Maryam Mirzaei$^{1}$, 
Yafei Ou$^{1}$, and Mahdi Tavakoli$^{1,2}$, \emph{Senior Member, IEEE}
\thanks{Manuscript received: August, 10, 2024; Revised November, 5, 2024; Accepted January, 19, 2025.}
\thanks{This paper was recommended for publication by Editor Pietro Valdastri upon evaluation of the Associate Editor and Reviewers' comments.
This research was supported by the Canada Foundation for Innovation (CFI), the Natural Sciences and Engineering Research Council (NSERC) of Canada, the Canadian Institutes of Health Research (CIHR), and Alberta Innovates. (Corresponding author: Sadra Zargarzadeh)} 
\thanks{$^{1}$Sadra Zargarzadeh, Maryam Mirzaei, Yafei Ou, and Mahdi Tavakoli are with the Department of Electrical and Computer Engineering, University of Alberta, Edmonton, AB, Canada. \{{\tt\footnotesize sadra.zar, mmirzaei, yafei.ou, mahdi.tavakoli\}@ualberta.ca}}%
\thanks{$^{2}$Mahdi Tavakoli is with the Department of Biomedical Engineering, University of Alberta, Edmonton, AB, Canada.}%
\thanks{Digital Object Identifier (DOI): see top of this page.}
}

%
%

\markboth{IEEE Robotics and Automation Letters. Preprint Version. Accepted January, 2025}
{Zargarzadeh \MakeLowercase{\textit{et al.}}: From Decision to Action in Surgical Autonomy}

%



\makeatletter

\newcommand*{\rom}[1]{\expandafter\@slowromancap\romannumeral #1@}
\newcommand{\linebreakand}{%
  \end{@IEEEauthorhalign}
  \hfill\mbox{}\par
  \mbox{}\hfill\begin{@IEEEauthorhalign}
}
\makeatother


\maketitle

\begin{tikzpicture}[remember picture, overlay]
\node [align=left, xshift=10.5cm, yshift=-1.6cm] at (current page.north west) 
{
\begin{minipage}{17.5cm} 
\scriptsize
\copyright 2025 IEEE. Personal use of this material is permitted. Permission from IEEE must be obtained for all other uses, in any current or future media, including reprinting/republishing this material for advertising or promotional purposes, creating new collective works, for resale or redistribution to servers or lists, or reuse of any copyrighted component of this work in other works.
DOI: 10.1109/LRA.2025.3535184
\end{minipage}
};
\end{tikzpicture}

\begin{abstract}
The rise of Large Language Models (LLMs) has impacted research in robotics and automation. While progress has been made in integrating LLMs into general robotics tasks, a noticeable void persists in their adoption in more specific domains such as surgery, where critical factors such as reasoning, explainability, and safety are paramount. Achieving autonomy in robotic surgery, which entails the ability to reason and adapt to changes in the environment, remains a significant challenge. In this work, we propose a multi-modal LLM integration in robot-assisted surgery for autonomous blood suction. The reasoning and prioritization are delegated to the higher-level task-planning LLM, and the motion planning and execution are handled by the lower-level deep reinforcement learning model, creating a distributed agency between the two components. As surgical operations are highly dynamic and may encounter unforeseen circumstances, blood clots and active bleeding were introduced to influence decision-making. Results showed that using a multi-modal LLM as a higher-level reasoning unit can account for these surgical complexities to achieve a level of reasoning previously unattainable in robot-assisted surgeries. These findings demonstrate the potential of multi-modal LLMs to significantly enhance contextual understanding and decision-making in robotic-assisted surgeries, marking a step toward autonomous surgical systems.
\end{abstract}

\begin{IEEEkeywords}
Medical robots and systems, multi-modal large language models, surgical robot, planning, laparoscopy.
\end{IEEEkeywords}

%
\IEEEpeerreviewmaketitle

\section{Introduction}
%
%
%
%
\IEEEPARstart{R}{obot}-assisted surgery (RAS) has enormously changed the way many surgeons operate. Surgical robots can enhance accuracy and dexterity, provide better anatomical access, and minimize invasiveness, surgery time, and the need for revision surgery \cite{taylor2016medical}. With the development of surgical robots and the da Vinci Research Kit (dVRK) \cite{kazanzides2014open}, along with realistic surgical simulation environments\cite{ou2024realistic, scheikl2023lapgym, xu2021surrol}, the automation of surgical sub-tasks such as tissue retraction \cite{scheikl2022sim}, suturing \cite{sen2016automating}, endoscopic camera control \cite{ou2023robot}, cutting \cite{nguyen2019manipulating}, and body fluid removal \cite{ou2024autonomous}, has been an area of research in the past few years. These are the building blocks of surgeries that form the foundation for enhancing bottom-up surgical autonomy \cite{lalys2014surgical, tagliabue2020soft}, and automating these commonly faced sub-tasks provides the basic robot skills necessary for reaching a more advanced level of autonomy, including the ability to reason and plan tasks.

\begin{figure}
\centering
\includegraphics[width=3.5in]{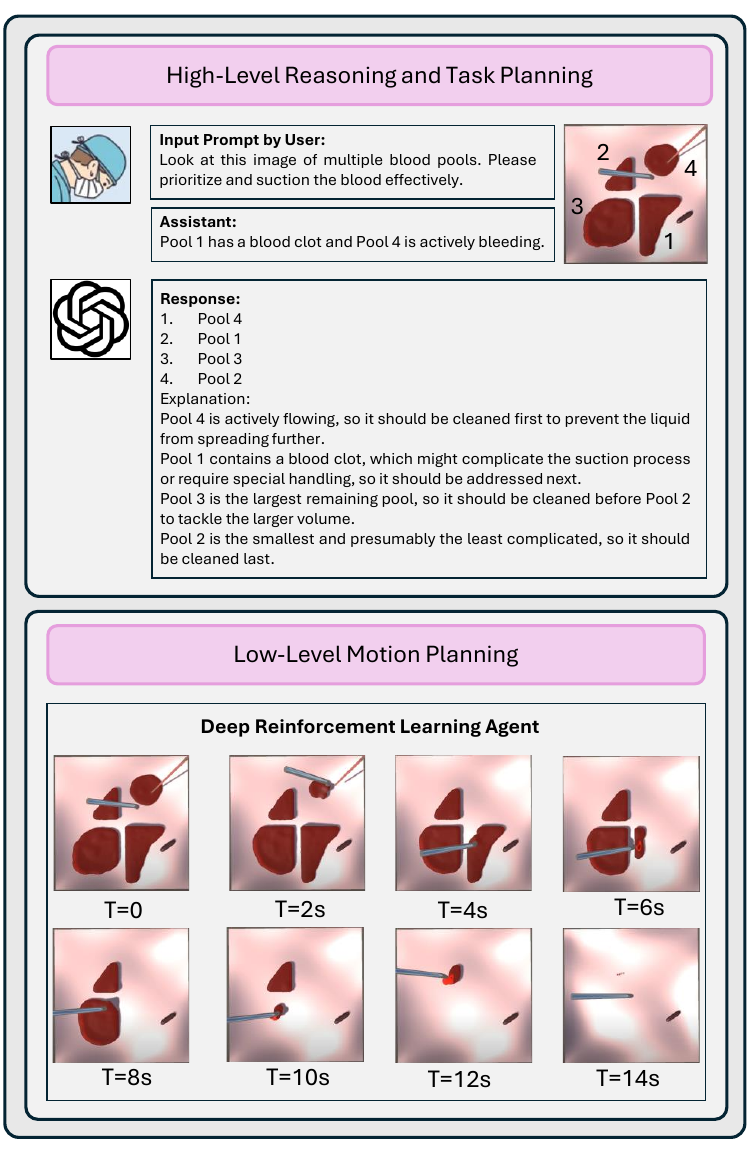}
\caption{\small {The high-level task reasoning and planning for the blood suction task is performed by the LLM, and the low-level motion planning and execution is done by the DRL agent.}}
\label{Figure 1}
\vspace{-15pt} 
\end{figure}

Autonomous execution of surgical sub-tasks is typically learned through model-based methods \cite{9561624} or data-driven approaches like deep reinforcement learning (DRL) \cite{ou2023sim} and imitation learning \cite{su2021toward, hu2023autonomous}. However, these methods often lead to behavior that lacks human interpretability, explainability, and adaptability in decision-making. They prioritize maximizing cumulative rewards and efficient exploration of the state space but do not implicitly account for the risks of the actions taken, thus failing to assure safety standards \cite{pore2021safe, garcia2015comprehensive, fan2024learn}. Moreover, DRL and imitation learning struggle with dynamic adaptability, particularly in scenarios where surgical conditions deviate from the norm due to unexpected patient anatomy or sudden complications such as bleeding. This inability to reason and act in unforeseen circumstances underscores the need for a reasoning framework that can adapt to challenges, ensuring efficacy and transparent decision-making in autonomous surgical tasks. As surgical sub-tasks involve physical interaction with patients, any automation and decision-making by the robot must be clear and understandable by the operator, resembling human-like behavior to ensure safety and reliability.

Autonomous surgical planning in robot-assisted sub-tasks demands a human-like reasoning unit capable of pre-operatively planning task execution and intraoperatively modifying the plan to accommodate unforeseen circumstances. This approach would enhance the explainability of the robot's decisions and minimize associated risks. Large Language Models (LLMs), trained on vast amounts of text data, have revolutionized natural language understanding and have been adopted in various domains beyond NLP, such as planning and interaction for robots. They can serve as a high-level reasoning unit, breaking down given commands into smaller subtasks to be executed by the robot's lower-level systems, including motion planning and control.

Integrating LLMs into robots poses challenges due to their struggle with real-world complexities. While LLMs offer general knowledge and expertise, they lack a connection to tangible reality, leading to errors and potentially unsafe recommendations. Extracting information from LLMs for robots requires balancing theoretical knowledge with practical understanding to navigate dynamic environments and facilitate effective human-robot interaction. An agent needs to comprehend semantic aspects of the world, the range of available skills, how these skills influence the environment, and how changes in the world map back to language \cite{huang2022inner}. The integration of robotics and LLMs, as introduced through Google's SayCan \cite{ahn2022can}, PaLM-E \cite{driess2023palm}, and more recent methods \cite{liu2023llm, kannan2023smart, wu2023tidybot, singh2023progprompt, vemprala2023chatgpt, long2023robollm, arenas2023prompt, dai2023think}, presents immense opportunities for exploration in domains such as surgical robotics.

By leveraging multi-modal LLMs in a zero-shot manner, in this work we aim to surmount the limitations of current autonomous systems, introducing a level of reasoning and adaptability previously unattainable in robot-assisted surgeries. Integrating images with text allows the multi-modal LLM to capture important visual nuances, such as spatial relationships and the presence of surgical tools near blood pools, which may not be fully conveyed through text alone. This integration is pioneering, as it combines the theoretical knowledge of LLMs with the practical demands of surgical environments as identified and communicated by the medical staff. We propose an LLM-powered framework capable of high-level reasoning, mid-level motion planning, and execution for autonomous blood suctioning in robot-assisted surgeries. The reasoning and prioritization responsibility is delegated to the higher-level task planning LLM, and the motion planning and execution are delegated to the DRL model, leading to a distributed agency between the two components. Expanding on the foundation laid with autonomous blood suctioning, our vision extends beyond this surgical subtask and sets a precedent for the future of autonomous systems in the surgical field.

\begin{figure}
\centering
\includegraphics[width=3.5in]{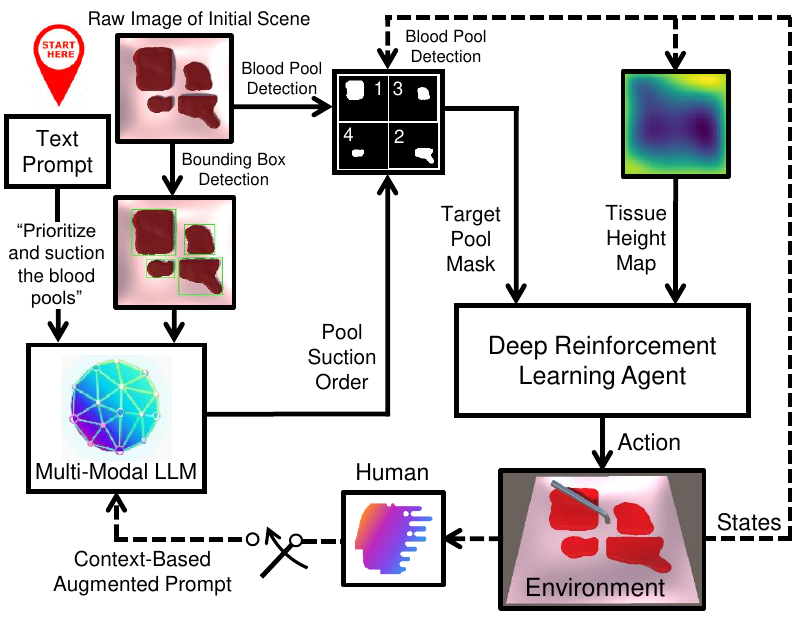}
\caption{\small {System architecture.}}
\label{Figure 2}
\end{figure}

The main contributions of this work are as follows:
\begin{enumerate}
    \item We propose an LLM-powered framework for autonomous robot-assisted blood suction, where task reasoning and planning are managed by the LLM, while motion planning and execution are handled by a DRL-trained agent.
    \item We compare the performance of LLM reasoning to random reasoning and no reasoning modules in terms of blood removal time and tool movement.
    \item We analyze how augmenting the prompts with context and expert-defined guidelines influences the reasoning capabilities of the LLM in zero-shot prompting. A user study is also conducted to assess the similarity to human-like behavior across the three modules.
\end{enumerate}

The paper is structured as follows. Section \rom{2} reviews the integration of LLMs in robotics and surgery and their impact on these fields. Section \rom{3} introduces the methods used, including the system architecture, multi-modal LLMs, prompt augmentation, the simulation environment, and the deep reinforcement learning module. Section \rom{4} presents and discusses the experiments and results. Section \rom{5} outlines the limitations and future work, while Section \rom{6} concludes the paper.

%

\begin{figure*}[t!]
\begin{center}
\centerline{\includegraphics[width=6.8in]{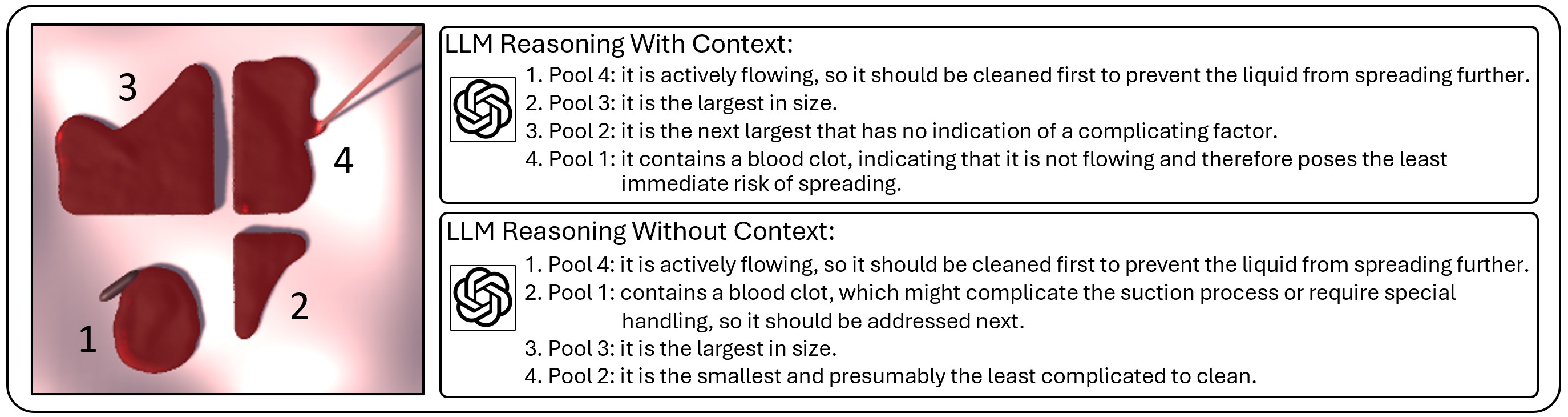}}
\caption{\small {An example of LLM reasoning with (LRWC) and without (LRWOC) context-based prompt augmentation. The guideline provided to the LLM is as context is as follows: Address active bleeding first, consider pool size next, and address the blood clot pool last, as coagulation ensures that flow in this pool has ceased and will not propagate further.}}
\label{Figure 3}
\end{center}
\vspace{-20pt} 
\end{figure*}

\section{Related Work}
\subsection{Large Language Models in Robotics}
In a pioneering work in LLM-Robotics integration, Google introduces SayCan \cite{ahn2022can}, a method that aims to extract knowledge stored within LLMs for physically grounded tasks. The LLM, `Say', breaks down instructions into subtasks and evaluates each skill's contribution likelihood. Affordance functions, `Can', assess each skill's success potential in the environment. This approach combines these factors to determine the effectiveness of each skill in fulfilling the instruction. In their later work, PaLM-E \cite{driess2023palm}, they propose an embodied language model that processes multi-modal sentences, blending visual, continuous state estimation, and textual inputs, leading to the integration of real-world continuous sensor data into language models and creating a direct connection between words and sensory experiences.

While these recent works have shown great advancement in the integration of LLMs into robotics, they mainly focus on general tasks such as object manipulation. More specific application domains, such as robot-assisted surgery, pose new and important challenges that need to be addressed to ensure the safety of patients in surgical operations. For instance, despite advancements, existing LLM applications have yet to fully tackle the real-time adaptation and decision-making required in dynamic surgical settings. Nevertheless, the success of LLMs in broader robotics tasks can lay a foundational understanding that is crucial when approaching more specialized domains such as robot-assisted surgery where data is scarce.


\subsection{Large Language Models in Surgery and Surgical Planning}
LLMs can assist in surgical planning by analyzing vast amounts of medical literature, patient data, and clinical guidelines to suggest the most appropriate surgical approaches. This includes evaluating the risks and benefits of different surgical options and considering patient-specific factors such as age and previous surgical history and can be used in various surgeries such as joint arthroplasty \cite{cheng2023potentials}, oral and maxillofacial surgery \cite{puladi2023impact}, and orthopedics \cite{chatterjee2023chatgpt}. It has the ability to process and generate complex language-based instructions, enabling bidirectional communication with the medical team in real-time in an intuitive, fast, and natural way. This capability is essential for improved decision-making in dynamic surgical environments, where rapid and accurate communication is crucial for patient safety and successful outcomes \cite{cheng2023potential}.

Although an initial step towards integrating LLMs in robotic surgery has been made in SUIFA \cite{moghani2024sufia} through tasks such as needle lift and shunt insertion, the current gaps in the literature include the adoption of LLMs as reasoning units in the planning process of surgical tasks where decision-making is crucial. In this work, we investigate the reasoning capability of LLMs in prioritizing the suction of multiple blood pools under different circumstances and integrate them into robot-assisted surgery for autonomous blood suctioning.

\section{Methods}
\subsection{Blood Suctioning Task and System Architecture}
Blood is among the most frequent types of fluids encountered in surgical settings, as bleeding is a common and sometimes unpredictable occurrence during operations. Surgeons typically need to address bleeding promptly by clearing the area and pinpointing its source before proceeding with other tasks. Suction of the blood with the proper tool becomes necessary to extract blood from the surgical site. Consequently, this task is indispensable and often consumes significant time and effort, and automating this process reasonably and safely can alleviate the burden on surgeons. In a dynamic environment, such as the human body, fluids such as blood move around, forming pools with different conditions. For instance, characteristics such as active bleeding, blood clots, variations in blood pool sizes, and the closeness of blood pools to critical organs or surgical instruments affect the priority with which they need to be suctioned.

As shown in Fig. 1, an image of the tissue scene containing multiple blood pools is annotated with bounding boxes around each pool and passed to the multi-modal large language model, along with the text prompt, ``Look at this image of multiple blood pools. Please prioritize and suction the blood effectively." The proposed system architecture, illustrated in Fig. 2, depicts how the multi-modal LLM, as the high-level reasoning unit, uses its reasoning capability to prioritize the order in which the blood pools need to be suctioned and informs the masking sensor accordingly. The pools are masked in the order that they need to be suctioned and then fed into the trained DRL agent, which acts as the lower-level motion planner,  along with the tissue height map, and leads to an action taken by the suction tool. In complex scenarios where additional information is needed for the LLM to prioritize suction, human input can augment the initial prompt with context.

\subsection{Multi-Modal Large Language Models}
Our methodology leverages multi-modal LLMs to allow for information from diverse modalities, including text and images. In multi-modal LLMs, textual data undergoes processing through a standard language model architecture. Feature embeddings extracted from image inputs are concatenated with the textual embeddings, yielding a multi-input representation. This fused representation is then fed into a multi-layer neural network, facilitating joint learning across modalities. In this work, we employed the pre-trained OpenAI GPT-4V model \cite{achiam2023gpt} for image understanding and reasoning.

As shown in Fig.~\ref{Figure 1} and explained in the system architecture section, the multi-modal LLM accepts a text prompt and an image of the labeled blood pools. The prompt outlines essential details about blood pools, including the presence of clots, and signs of active bleeding. Each of these aspects is crucial in deciding the urgency of suctioning blood pools during medical procedures. Size indicates the amount of blood, while clots indicate coagulation and cessation of blood flow. Active bleeding is a significant signal that could suggest the potential for blood to spread more extensively. By incorporating these factors into the prompt, we equip the multi-modal LLM with essential context for making informed decisions regarding blood pool suctioning. The LLM conducts reasoning based on the image and the prompt, yielding the proper priority for suctioning the blood, and explains why this priority is chosen in a zero-shot manner. Our zero-shot approach allows the LLM to generate relevant responses for each prompt without specific training examples, relying on its general understanding of language semantics and prompt cues to address previously unseen conditions effectively.

In scenarios where we encounter a combination of factors concerning blood pools simultaneously, such as when both active bleeding and a blood clot exist in two of the pools, the LLM may lack consistency in reasoning due to the lack of training on medical data. To address this, we activate the switch as seen in Fig.~\ref{Figure 2} and exemplified in Fig.~\ref{Figure 3}, and augment the prompt with additional contexts providing a guideline for the model in generating a consistent priority in complex scenarios. This approach remains in a zero-shot manner as we refrain from furnishing the model with specific examples. Instead, we enhance the prompt through context augmentation as part of a prompt engineering process.

\subsection{Blood Suction Environment and Mask Sensor Mechanism}
In our recent work \cite{ou2024autonomous}, a blood suction simulation environment for RL was built using position-based fluids (PBF) based on Nvidia PhysX 5, Unity, and Unity's ML-Agents toolkit. PBF is an approach that represents fluids using a large number of small particles that interact with each other. PhysX is a real-time physics engine with GPU optimization, which allows for PBF simulation. With PhysX 5 as the low-level physics engine, the main simulation environment was built in Unity. In this model, blood is simulated as particles influenced by forces like gravity and suction, with a spherical cone-shaped force field applied to particles near the suction tool. The force decays with distance to realistically simulate suction, removing particles once they reach a specified height threshold. We leverage the existing simulation environment and further build upon it in this work.

\begin{figure}[t!]
\begin{center}
\centerline{\includegraphics[width=3.4in]{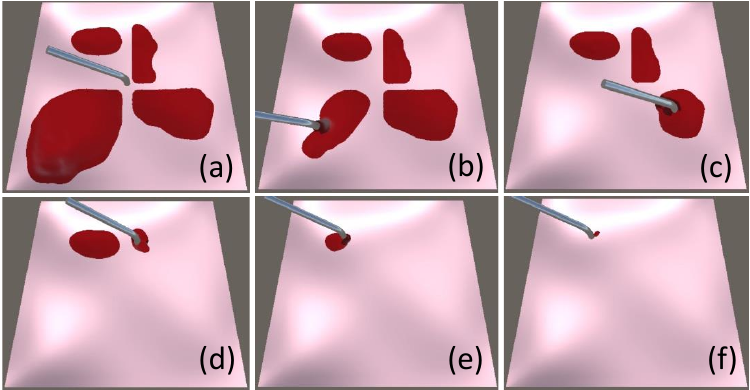}}
\caption{\small {Simulation Environment 1. The LLM reasoning prioritizes suctioning the pools based on their size in the absence of surgical complexities such as active bleeding and blood clots as seen in (a)-(f).}}
\label{Figure 4}
\end{center}
\vspace{-30pt} 
\end{figure}

The simulation environment consists of a randomly generated tissue that contains the blood, the simulated blood, and a suction tool. We simulated a fixed amount of blood (4000 particles) through PBF and added suction force to each particle within a suction range to simulate the effect of suction. Particles that are suctioned close enough to the suction tool will be removed and marked as inactive. To introduce randomness in the shape of the tissue, Bezier surfaces with random control points were used to generate random shapes.
The Bezier surfaces are represented by
\begin{equation}
    \mathcal{S}(u, v) = \sum_{i=0}^{n} \sum_{j=0}^{m} P_{i,j} \cdot B_{n,i}(u) \cdot B_{m,j}(v),\ 0\leq u, v\leq1
\end{equation}
where \(P_{i,j}\) are the control points, and \(B_{n,i}(u)\) and \(B_{m,j}(v)\) are the Bernstein basis functions defined by
\begin{equation}
    \\B_{n,i}(x) = \frac{n!}{i! \cdot (n - i)!} \cdot x^i \cdot (1 - x)^{n - i}.\
\end{equation}

\begin{figure*}[t!]
\centering
\begin{subfigure}[b]{0.5\textwidth}
    \centering
    \caption{Environment 1 (No blood clot or active bleeding)}
    \includegraphics[width=3.3in]{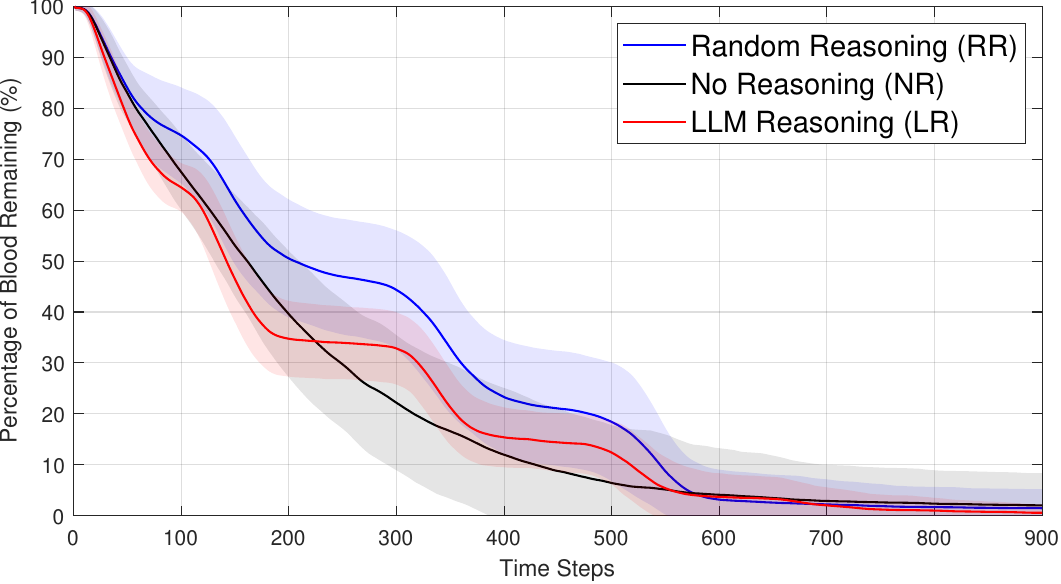}
    \label{fig:1}
\end{subfigure}%
\begin{subfigure}[b]{0.5\textwidth}
    \centering
    \caption{Environment 2 (Only active bleeding)}
    \includegraphics[width=3.3in]{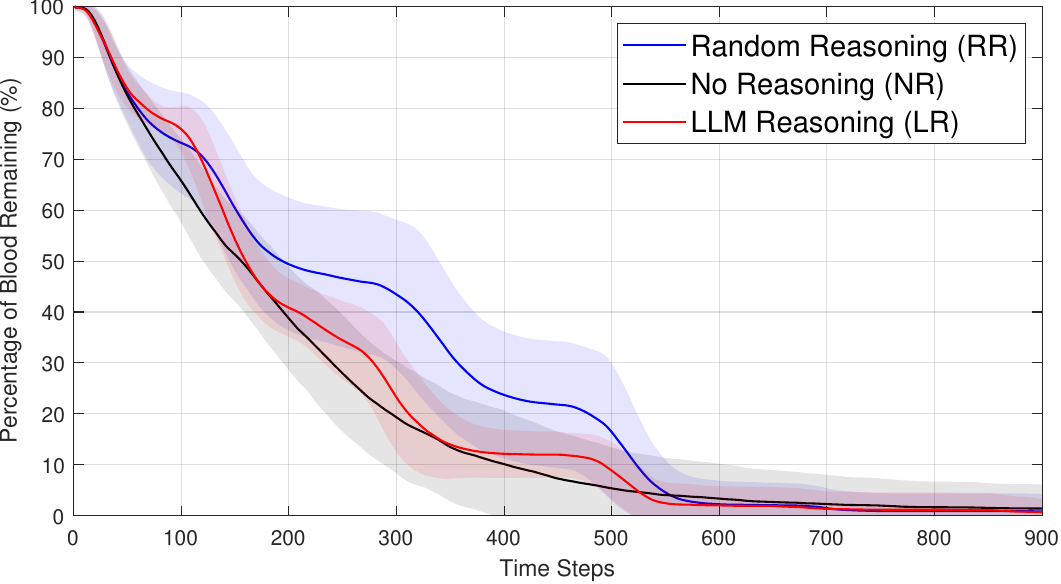}
    \label{fig:2}
\end{subfigure}
\begin{subfigure}[b]{0.5\textwidth}
    \centering
    \caption{Environment 3 (Only blood clot)}
    \includegraphics[width=3.3in]{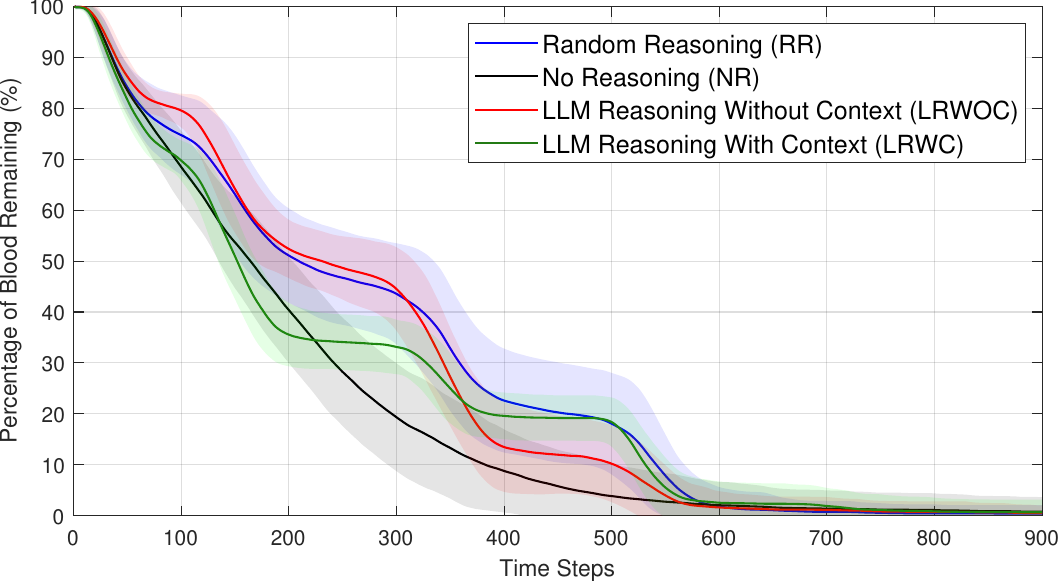}
    \label{fig:3}
\end{subfigure}%
\begin{subfigure}[b]{0.5\textwidth}
    \centering
    \caption{Environment 4 (Both active bleeding and blood clot)}
    \includegraphics[width=3.3in]{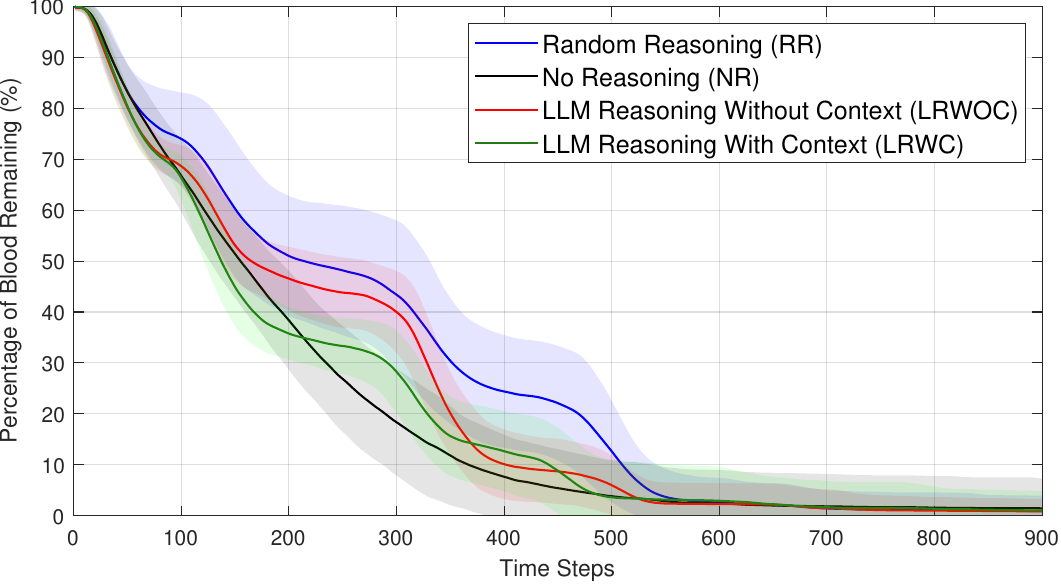}
    \label{fig:4}
\end{subfigure}
\caption{\small {Progression in blood suction in the four environments.}}
\label{Figure 5}
\end{figure*}

In this work, several features are added to the simulation environment, including a module that would allow blood to continuously add to a random pool representing active bleeding, the addition of a capsule-shaped object representing a blood clot that would be randomly positioned on the tissue, and the ability to generate multiple independent blood pools. We also developed a blood pool detector algorithm that takes in the raw image of the scene along with the suction orders from the multi-modal LLM and outputs the mask of the target blood pool based on the priority reasoned by the LLM. This mechanism would allow the agent to plan the suction motion of only the target pool and would blind the agent to other existing pools unless commanded otherwise by the LLM. 

We developed four simulation environments. In Environment 1, as can be seen in Fig. \ref{Figure 4}, four independent pools are randomly generated without the presence of a blood clot or active bleeding. In Environment 2, active bleeding occurs randomly in one of the pools for a fixed time interval, with no sign of a blood clot. Environment 3 features the introduction of a capsule-shaped object representing a blood clot, randomly generated near one of the pools, indicating coagulation and the cessation of potential bleeding. Finally, Environment 4 includes a pool actively bleeding, while another contains a blood clot. To test the realism of our simulation environments, visual comparisons with real surgical environments were conducted, ensuring our model visually reflects the dynamics of blood flow and bleeding during real surgical procedures.

\subsection{Motion Planning Using Deep Reinforcement Learning}

In our recent work \cite{ou2024autonomous}, an RL agent for completing autonomous blood suction was obtained. To train this agent, we used the following reward function, which consists of a reward for the amount of blood removed during each step, an extra terminal reward for removing all blood, and an action penalty for tool movements. The number of particles being removed during each step is used to determine the amount of blood being suctioned out.
\begin{equation}
    r(s_t,a_t,s_{t+1}) = N_p^{t} - N_p^{t+1} + C_1\, \delta(N_p^{t+1}) - C_2 \| a_t \|
\end{equation}
In the above equation, $t$ is the time step, $N_p^t$ is the number of active particles, $\delta(N_p^{t+1})$ denotes whether there are active particles remaining, and $\| a_t \|$ is the norm of the actions. 
The weighting factors $C_1=5$ and $C_2=0.02$ were chosen to balance task efficiency and control stability, emphasizing blood removal while discouraging excessive motion. Specifically, $C_1$ is set to a higher value to prioritize the reward for full blood clearance, encouraging the agent to complete the task efficiently, while $C_2$ is relatively small, penalizing movements without restricting necessary adjustments.
The observation includes the tissue height map, the binary image mask of the blood (stacked with 3 from previous steps), and the suction tool location (stacked with 4 from previous steps). The binary image mask of the blood is synthesized from the current positions of all active particles in the blood.

\section{Experiments and Results}
We investigate four reasoning modules in our experiments. When the LLM, as the high-level reasoning unit, reasons the sequence for suctioning blood pools without any additional context, it is termed LLM Reasoning Without Context (LRWOC). If additional context is provided by the assistant, leading to an augmented prompt for the LLM, we call this LLM Reasoning With Context (LRWC). When the DRL agent receives a randomly generated order based on a random permutation of the number of blood pools, this is known as Random Reasoning (RR). If the DRL agent proceeds to suction blood pools solely based on its reward function without input from a higher-level unit, it is termed No Reasoning (NR).

\subsection{Comparison of LLM Reasoning with Random Reasoning and No Reasoning}
To evaluate the performance of the RR, NR, LRWOC, and LRWC modules, we simulated the blood suction task across 400 distinct scenes (100 scenes per environment). Fig. \ref{Figure 5} illustrates the blood remaining percentage over time, providing a comparative analysis of the different reasoning modules across different environments. Additionally, Table I presents key metrics such as the mean and standard deviations of the time to suction the actively bleeding pool in Environments 2 and 4 ($T_{AB}$), the time to suction 50\% ($T_{50}$) and 95\% ($T_{95}$) of the blood, and total tool path length (TTPL).

In Environment 1 (Fig. \ref{Figure 5}a), the LLM reasoning module reasons that larger volumes of blood must be addressed first and prioritizes suctioning the pools based on their size, resulting in a more rapid initial suction compared to the other modules leading to a faster average $T_{50}$. The NR module shows a gradual decrease in blood remaining by suctioning parts of pools as it moves between them, while the LR and RR modules prefer to suction one pool before moving to the next, resulting in intervals where the slope decreases indicating movement between pools.

\begin{figure}[t]
\begin{center}
\centerline{\includegraphics[width=3.3in]{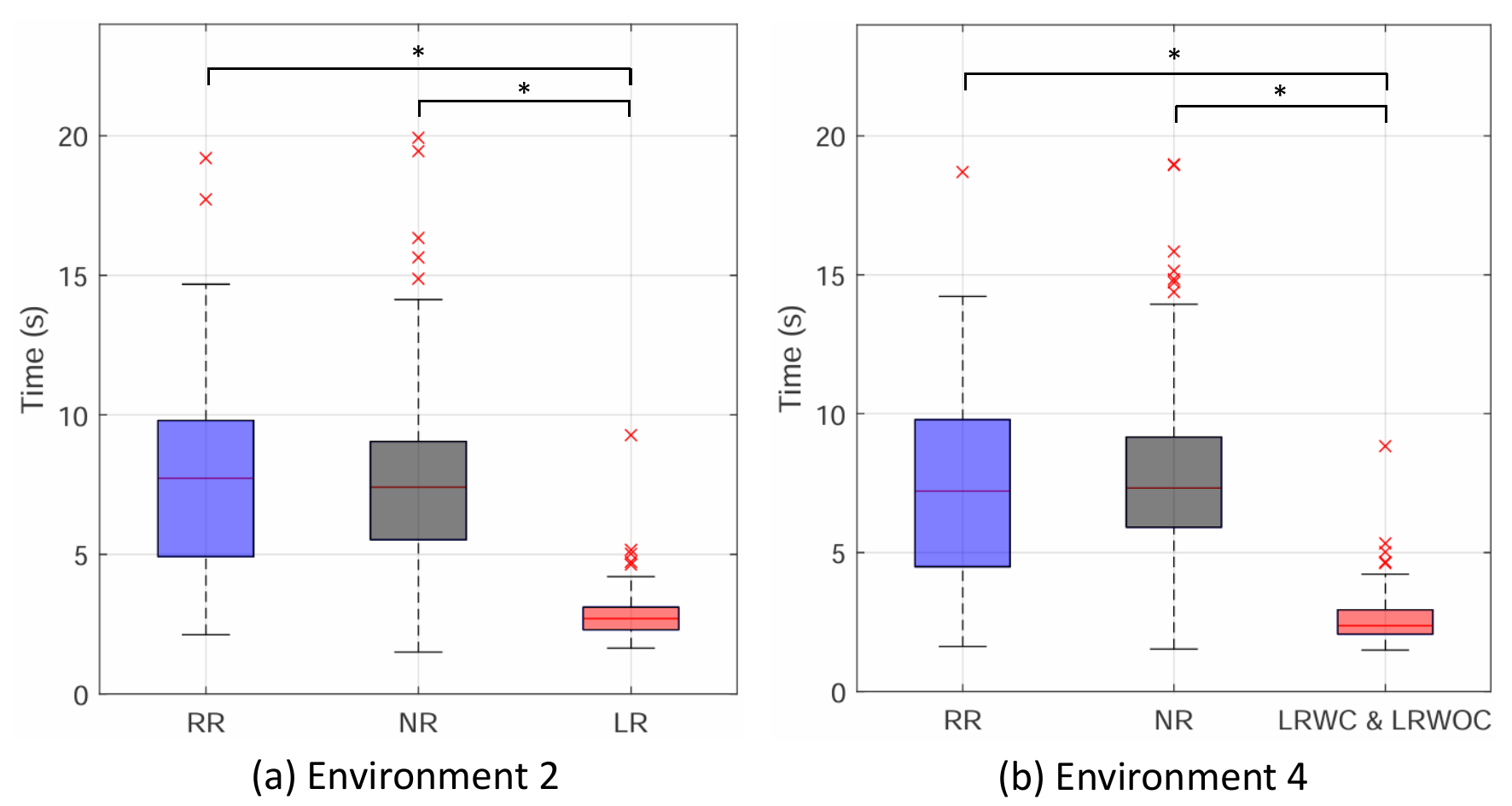}}
\caption{\small {Time to suction the actively bleeding pool ($T_{AB}$) in Environments 2 and 4 by different reasoning modules.}}
\label{Figure 6}
\end{center}
\vspace{-25pt} 
\end{figure}

In Environment 2 (Fig. \ref{Figure 5}b), where active bleeding is present in one of the pools, the LLM reasoning module gives priority to suctioning the pool with active bleeding first, even if it is smaller than others. This approach results in a faster $T_{AB}$. After addressing the pool with active bleeding, the LLM then proceeds to prioritize suctioning the pools based on their size, similar to the strategy observed in Environment 1. Providing additional instructions as the augmented prompt did not change the LLM reasoning in Environments 1 and 2 and the focus remained on prioritizing active bleeding, followed by pool size. This resulted in the same LRWOC and LRWC, denoted as LR.

In Environment 3 (Fig. \ref{Figure 5}c), the presence of a blood clot adds complexity to suctioning. LRWOC tends to prioritize the suction of the pool with the blood clot potentially due to its perceived complexity. However, the user can define rules for the LLM to follow, tailoring its reasoning to specific conditions. For this environment, we establish the following rule: 1) Address active bleeding first, 2) Consider pool size next, and 3) Address the blood clot pool last, as coagulation ensures that flow in this pool has ceased and will not propagate further. An example of this is shown in Fig.~\ref{Figure 3}. This context leads to a faster average $T_{50}$ in LRWC, prioritizing blood pool size after active bleeding. In Environment 4 (Fig. \ref{Figure 5}d), both LRWOC and LRWC modules start by suctioning actively bleeding pools. LRWC prioritizes larger pools next and saves the pool with the blood clot for last, resulting in a faster $T_{50}$ compared to LRWOC, which targets the pool with the blood clot before proceeding based on pool size.

\begin{table}[]
\centering
\renewcommand{\arraystretch}{1.3}
\centering
\caption{\small {Key metrics in comparison of reasoning modules expressed as mean$\pm$std (Time step = 0.02 seconds).}}
\resizebox{\columnwidth}{!}{%
\begin{tabular}{cccccc}
\hline
\multicolumn{2}{c}{}                                      & $T_{AB}$  & $T_{50}$  & $T_{95}$   & TTPL \\ \hline
\multicolumn{1}{c|}{\multirow{3}{*}{Environment 1}} & RR   & -- & 225$\pm$82   & 570$\pm$128 & 33.9$\pm$3.2          \\
\multicolumn{1}{c|}{}                               & NR  & --  & 164$\pm$38   & 557$\pm$110 & 32.4$\pm$4.1          \\
\multicolumn{1}{c|}{}                               & \textbf{LR} & --   & \textbf{145$\pm$30}   & \textbf{554$\pm$101} & \textbf{32.5$\pm$2.8}          \\ \hline
\multicolumn{1}{c|}{\multirow{3}{*}{Environment 2}} & RR   & 363$\pm$176 & 223$\pm$89   & 573$\pm$130 & 33.3$\pm$3.4          \\
\multicolumn{1}{c|}{}                               & NR  & 390$\pm$174  & 154$\pm$37   & 514$\pm$106 & 31.5$\pm$3.9          \\
\multicolumn{1}{c|}{}                               & \textbf{LR}  & \textbf{128$\pm$50}  & \textbf{161$\pm$28}   & \textbf{523$\pm$95} & \textbf{33$\pm$2.8}            \\ \hline
\multicolumn{1}{c|}{\multirow{4}{*}{Environment 3}} & RR   & -- & 222$\pm$71   & 563$\pm$124 & 32.4$\pm$2.9          \\
\multicolumn{1}{c|}{}                               & NR  & --  & 166$\pm$34    & 466$\pm$98 & 31.2$\pm$4.2          \\
\multicolumn{1}{c|}{}                               & \textbf{LRWOC} & -- & 239$\pm$58   & 541$\pm$103 & 32.4$\pm$3.0          \\
\multicolumn{1}{c|}{}                               & \textbf{LRWC} & -- & \textbf{151$\pm$23}   & \textbf{553$\pm$100} & \textbf{32.7$\pm$2.9}          \\ \hline
\multicolumn{1}{c|}{\multirow{4}{*}{Environment 4}} & RR  & 359$\pm$178  & 227$\pm$86    & 535$\pm$120 & 33.7$\pm$3.2          \\
\multicolumn{1}{c|}{}                               & NR  & 393$\pm$172  & 155$\pm$34   & 462$\pm$104 & 30.4$\pm$3.8          \\
\multicolumn{1}{c|}{}                               & \textbf{LRWOC} & \textbf{130$\pm$49} & 179$\pm$55   & 509$\pm$93 & 33.7$\pm$2.9          \\
\multicolumn{1}{c|}{}                               & \textbf{LRWC} & \textbf{130$\pm$49}  & \textbf{137$\pm$18}   & \textbf{470$\pm$90} & \textbf{32.9$\pm$2.8}          \\ \hline
\end{tabular}%
}
\vspace{-15pt}
\end{table}

Fig. \ref{Figure 6} illustrates the time to suction the actively bleeding pool ($T_{AB}$) in Environments 2 and 4 and shows statistically significant improvement (*) when LLM reasoning is used. Table I shows that the LRWC module results in smaller standard deviations across all metrics and environments, indicating more consistent performance. Although the NR module exhibited a marginally smaller average TTPL, its inability to reason and adapt to unforeseen circumstances makes it unreliable in highly dynamic surgical settings.

\subsection{Advantage of Multi-Modal LLMs}
To further demonstrate the advantage of the multi-modal LLM, we conducted an experiment to test its ability to capture contextual details that may not be explicitly provided in text. In this experiment, we presented the model with 10 images in which a surgical tool was positioned near one of the blood pools, using our original prompt without any mention of the tool in text. The multi-modal LLM correctly recognized the presence of the tool in 80\% of cases (8 out of 10) and incorporated this visual information into its decision-making process, placing this priority right after the active bleeding pool. Unlike hard-coded logic, which lacks flexibility in unforeseen scenarios, the LLM leverages context awareness and visual information to adapt to these conditions interpreting nuanced visual cues that a text-only input or a hard-coded logic might overlook.

\subsection{User Study on Closeness to Human-Like Behaviour}
To assess the similarity to human-like behavior exhibited by the RR, NR, and LRWOC modules during blood suction, a survey questionnaire was conducted involving ten participants. The participants, consisting of graduate students and senior researchers with no specialized medical backgrounds, were chosen to assess how well a non-expert-defined context aligns with broader, non-specialist perspectives in evaluating the behavior of the RR, NR, and LRWOC modules. A total of thirty-six videos were collected for the survey, divided among three reasoning modules (RR, NR, LRWOC), with each module contributing twelve videos from four environments (three videos per environment), showcasing the blood suctioning task. Participants were presented with pairs of videos and asked, ``If you were the human operator, which of the two videos shown below would you choose to suction the blood pools?". This forced choice questioning process was repeated to compare all combinations of LRWOC versus NR versus RR. 

A second user study survey involved collecting nine videos each of LRWOC and LRWC (in Environments 3 and 4), which were then presented to the participants with the same question. A human performance score was defined as the number of videos selected by the participants normalized by the total number of videos, as shown in Fig. \ref{Figure 7}. The objective of these surveys was to investigate whether LLM reasoning aligns more with actual human decision-making than random reasoning and no reasoning modules and also to assess how providing our user-defined context impacts the LLM's ability to mimic human decision-making. This research was approved by the University of Alberta's Research Ethics Board under approval ID Pro00139696.

A one-way analysis of variance (ANOVA) test was used in the first survey and a paired t-test in the second survey to establish statistical significance among reasoning modules, as shown in Fig. \ref{Figure 7}. The first study yielded a $p$-value\textless 0.001, showing that LRWOC had significantly more human-like blood suction behavior. The second study also resulted in a $p$-value\textless 0.001. The user study results, indicating a preference for the LRWC over the LRWOC module's decision-making, suggest that incorporating contextual understanding in robotic surgery could bridge the gap between automated procedures and the intuitive decision-making of humans.

Although results show a promising stride towards explainable surgical autonomy, a thorough evaluation process including clinical trials is essential to establish the efficacy and safety of LLM-powered robotic systems in real-world surgical applications. Additionally, the development of new training protocols for surgical teams on interacting with and overseeing LLM-enabled systems, along with intuitive interfaces for surgeons to interact with and override the system’s decisions when necessary, will be vital for adoption.

\section{Limitations and Future Work}
While the proposed method demonstrates the effectiveness of multi-modal LLMs in reasoning and decision-making within a simulated environment, several limitations and areas for future work remain, particularly for transitioning to real-world applications in surgical settings.

This study assumed that blood pools are separate and independent, simplifying interactions within the environment. Additionally, the current system operates below real-time performance due to the generation speed of OpenAI’s GPT-4V, which constrains its immediate applicability in time-sensitive clinical tasks. The physical experiments demonstrated in our previous work in blood suction \cite{ou2024autonomous} show the real-world execution of this task and hence were not the main focus of this study.

Simulation-based environments, while inherently limited compared to real-world settings, offer the critical advantage of encompassing a wide range of scenarios, including rare but pivotal situations that are difficult to consistently reproduce in physical setups. This simulation-first approach provides a foundation to test and refine the system’s decision-making capabilities across varied conditions, setting the groundwork for application in real surgical environments.

\begin{figure}[t]
\begin{center}
\centerline{\includegraphics[width=3in]{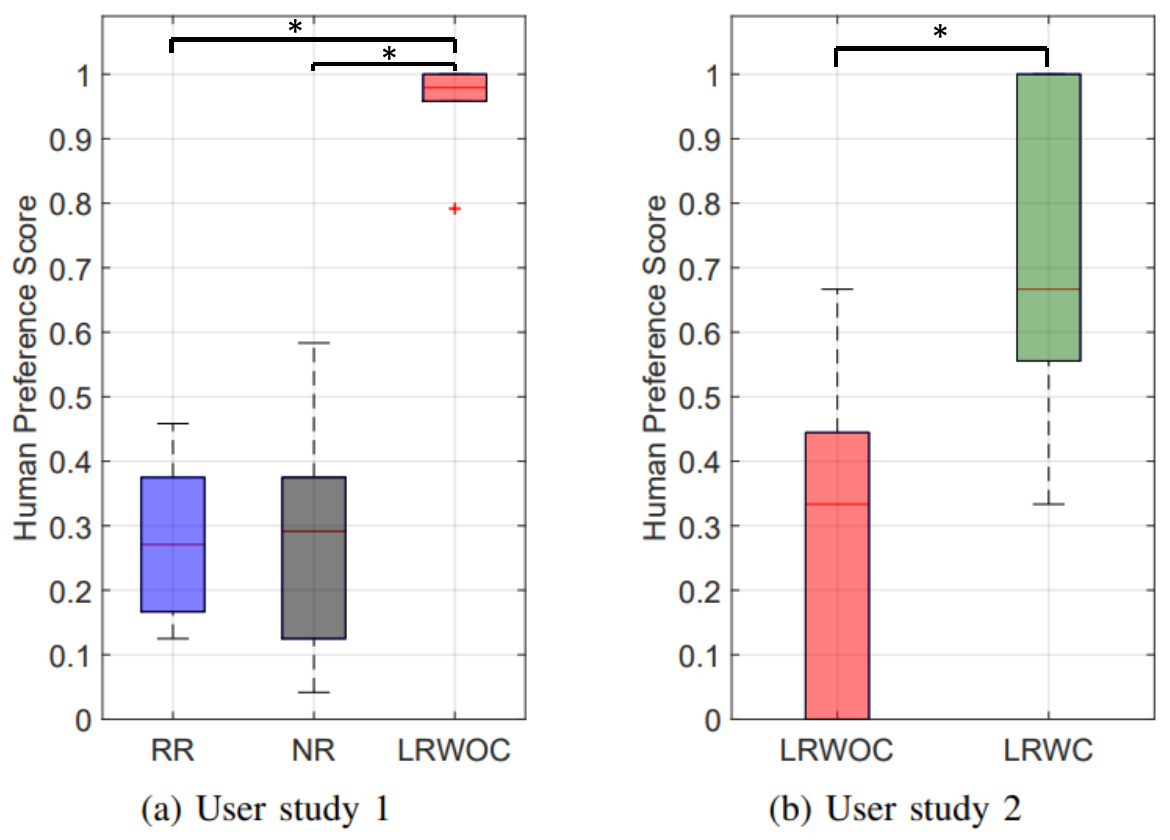}}
\caption{\small{Comparison on closeness to human-like behavior of
different reasoning modules.}}
\label{Figure 7}
\end{center}
\vspace{-27pt} 
\label{fig:composite}
\end{figure}

Applying this framework to real surgical contexts introduces practical challenges, such as accurate segmentation of blood from live camera images, real-time pose coordination of surgical instruments, and adapting to the dynamic and complex environment of an actual operating room. Future work will address these challenges by refining and validating the framework’s applicability in real surgical tasks, exploring predictive mechanisms and pre-emptive prompts to prioritize time-sensitive actions. To enhance real-time feasibility, model distillation, quantization, and smaller, task-specific models will be investigated, enabling the system to better align with the operational demands of clinical practice.

Ensuring the accuracy and safety of LLM-based decisions is another priority. LLMs are prone to hallucinations, which could impact decision reliability. To mitigate this, future iterations will integrate rule-based checks and domain-specific constraints, as well as feedback loops from medical experts, to improve the system’s robustness and decision accuracy in high-stakes environments. Gathering insights from medical professionals will also help assess the LLM’s alignment with expert-defined surgical priorities, refining its adaptability to clinical needs.

Safety in surgical tasks is essential, and the current framework will be expanded to address this by incorporating force-based thresholds to control tool speed and acceleration, collision detection, and Safe Reinforcement Learning (Safe RL) techniques, such as reward shaping and risk-sensitive policies, to enhance safe tool proximity to sensitive tissues. Additionally, to better handle dynamic scenarios, future work will explore mid-sequence task re-planning to adapt to external disturbances and refine LLM reasoning capabilities through reinforcement learning from human feedback, ultimately improving the framework’s scalability to other surgical and medical tasks.

\section{Conclusion}
In this study, we proposed a multi-modal LLM integration in robot-assisted surgery for autonomous blood suction and investigated how the addition of a high-level reasoning unit can influence decision-making and performance. Experiments were conducted to analyze LLM reasoning in comparison to random reasoning and no reasoning modules. Active bleeding and blood clots were introduced to influence decision-making as is also common in highly dynamic surgical settings. Results showed that the presence of a multi-modal LLM as a higher-level reasoning unit can account for these surgical complexities in decision-making and prioritization to achieve a level of reasoning and explainability previously unattainable in robot-assisted surgeries. The user study showed that incorporating contextual understanding in robotic surgery could bridge the gap between automated procedures and the intuitive decision-making of humans.

\ifCLASSOPTIONcaptionsoff
  \newpage
\fi



\bibliographystyle{IEEEtran}
\bibliography{References}
\end{document}